# A Differential Approach to Inference in Bayesian Networks

Adnan Darwiche
Computer Science Department
University of California, Los Angeles, Ca 90095
darwiche@cs.ucla.edu

## Abstract

We present a new approach for inference in Bayesian networks, which is mainly based on partial differentiation. According to this approach, one compiles a Bayesian network into a multivariate polynomial and then computes the partial derivatives of this polynomial with respect to each variable. We show that once such derivatives are made available, one can compute in constant-time answers to a large class of probabilistic queries, which are central to classical inference, parameter estimation, model validation and sensitivity analysis. We present a number of complexity results relating to the compilation of such polynomials and to the computation of their partial derivatives. We argue that the combined simplicity, comprehensiveness and computational complexity of the presented framework is unique among existing frameworks for inference in Bayesian networks.

## 1 Introduction

Consider the probability table on the left of Figure 1 and suppose that our goal is to compute probabilities of events with respect to this table, say $Pr(a)$, $Pr(a, \bar{b})$, etc.[1] We can do this in the usual way. To compute $Pr(\mathbf{e})$ we simply identify the rows in the table which are consistent with $\mathbf{e}$ and sum up their probabilities. Alternatively, we can *compile* the probability table as follows. We *parameterize* the table as shown on the right of Figure 1, where we introduce what we call *evidence indicators* into each row. We then add up all rows to generate the multivariate polynomial:

$$\mathcal{F}(\lambda_{\bar{a}}, \lambda_{\bar{b}}, \lambda_b, \lambda_a)$$
$$= .03\lambda_a\lambda_b + .27\lambda_a\lambda_{\bar{b}} + .56\lambda_{\bar{a}}\lambda_b + .14\lambda_{\bar{a}}\lambda_{\bar{b}},$$

which is depicted graphically in Figure 2.

| A | B | $\phi$ | A | B | $\phi$ |
|---|---|---|---|---|---|
| true | true | .03 | true | true | $\lambda_a \lambda_b .03$ |
| true | false | .27 | true | false | $\lambda_a \lambda_{\bar{b}} .27$ |
| false | true | .56 | false | true | $\lambda_{\bar{a}} \lambda_b .56$ |
| false | false | .14 | false | false | $\lambda_{\bar{a}} \lambda_{\bar{b}} .14$ |

Figure 1: Probability table and its parameterization.

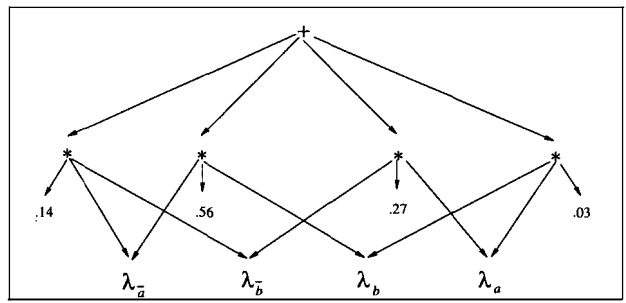

Figure 2: A probability-table compilation.

Now, given any evidence $\mathbf{e}$, we can compute the probability of $\mathbf{e}$ by a simple evaluation of polynomial $\mathcal{F}$. We consider each indicator $\lambda_x$. If $x$ is consistent with evidence $\mathbf{e}$, we set the value of variable $\lambda_x$ to 1. Otherwise, we set its value to 0. We then evaluate the polynomial $\mathcal{F}$, where its value is guaranteed to be the probability of evidence **e**. All we are doing here is using the indicators to select which rows of the probability table to add together. Figure 3 depicts some examples of evidence and the corresponding values of indicators. Figure 4 shows an example evaluation to compute the probability of $a$.

---

[1] We are using the standard notation: variables are denoted by upper–case letters ($A$) and their values by lower–case letters ($a$). Sets of variables are denoted by bold–face upper–case letters (**A**) and their instantiations are denoted by bold–face lower–case letters (**a**). For a variable $A$ with values true and false, we use $a$ to denote $A=$ true and $\bar{a}$ to denote $A=$ false.

| Evidence **e** | $\lambda_{\bar{a}}$ | $\lambda_{\bar{b}}$ | $\lambda_b$ | $\lambda_a$ |
|---|---|---|---|---|
| $a, b$ | 0 | 0 | 1 | 1 |
| $a, \bar{b}$ | 0 | 1 | 0 | 1 |
| $a$ | 0 | 1 | 1 | 1 |
| true | 1 | 1 | 1 | 1 |

Figure 3: Examples of evidence indicators.

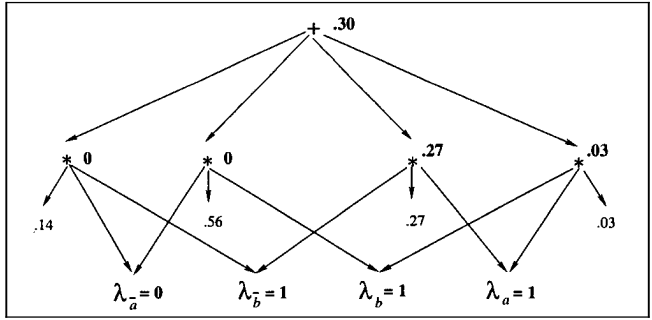

Figure 4: Evaluating a probability-table compilation under evidence $\mathbf{e} = a$.

The previous compilation technique can be applied to Bayesian networks, allowing one to generate factored polynomials that are not necessarily exponential in the number of network variables. We have promoted the compilation of Bayesian networks in earlier work, where we referred to polynomials such as $\mathcal{F}$ as Query-DAGs [6]. Our original motivation for compiling Bayesian networks was to simplify inference systems to the point where they could be implemented cost-effectively on a variety of software and hardware platforms. This paper is based on a number of new results, which make this notion of compilation much more interesting and useful than was originally conceived:

1. We show how to compile a Bayesian network into a polynomial that includes two types of variables: evidence indicators $\lambda_x$ and network parameters $\theta_\mathbf{f}$. Given a variable elimination order of with $w$ and length $n$, we show how to compile the polynomial in $O(n \exp(w))$ time using a simple variable elimination algorithm.

2. We show how to answer a large number of queries relating to classical inference [10, 13, 7, 21, 20], parameter estimation [19, 16], model validation [4] and sensitivity analysis [15, 2, 3] in constant-time once the partial derivatives of the compiled polynomial are computed.

3. We show the following on the complexity of computing the derivatives of such polynomials:

    (a) first partial derivatives, $\partial \mathcal{F}/\partial \lambda_x$ and $\partial \mathcal{F}/\partial \theta_\mathbf{f}$, can all be computed simultaneously in time linear in the size of polynomial $\mathcal{F}$, that is, in $O(n \exp(w))$ time.

    (b) second partial derivatives, such as $\partial^2 \mathcal{F}/\partial \lambda_x \partial \theta_\mathbf{f}$, can all be computed simultaneously in $O(n^2 \exp(w))$ time.

Following are examples of the queries that can be answered in constant time once these partial derivatives are computed:

1. the posterior marginal of any network variable $X$, $Pr(x \mid \mathbf{e})$;

2. the posterior marginal of any network family $\{X\} \cup \mathbf{U}$, $Pr(x, \mathbf{u} \mid \mathbf{e})$;

3. the sensitivity of $Pr(\mathbf{e})$ to change in any network parameter $\theta_\mathbf{f}$;

4. the probability of evidence **e** after having changed the value of some variable $E$ to $e$, $Pr(\mathbf{e} - E, e)$;[2]

5. the posterior marginal of some variable $E$ after having retracted evidence on $E$, $Pr(e \mid \mathbf{e} - E)$.

6. the posterior marginal of any pair of network variables $X$ and $Y$, $Pr(x, y \mid \mathbf{e})$;

7. the posterior marginal of any pair of network families $\mathbf{F}_1$ and $\mathbf{F}_2$, $Pr(\mathbf{f}_1, \mathbf{f}_2 \mid \mathbf{e})$;

8. the sensitivity of conditional probability $Pr(y \mid \mathbf{e})$ to a change in network parameter $\theta_\mathbf{f}$;

9. the amount of change to parameter $\theta_\mathbf{f}$ needed to ensure that $Pr(y \mid \mathbf{e}) \leq Pr(\bar{y} \mid \mathbf{e})$.

This paper is structured as follows. We introduce the polynomial representation of a Bayesian network in Section 2 and explicate some of its key properties. We then introduce the partial derivatives of this polynomial in Section 3 and present two key theorems that explicate their probabilistic semantics, followed by a number of corollaries showing how key probabilistic queries can be retrieved from such derivatives. Section 4 presents a simple variable-elimination algorithm for computing the polynomial representation of a Bayesian network while presenting structure-based guarantees on its time and space complexity. Section 5 presents an algorithm for evaluating the polynomial and for computing its partial derivatives efficiently. We finally close in Section 6 with some concluding remarks. Proofs of all theorems can be found in the full paper [5].

---

[2]We define this notation formally later.

## 2 Polynomial Representation of Bayesian Networks

Our goal in this section is to show how to represent a Bayesian network as a polynomial $\mathcal{F}$ and to explicate some of the key properties of this polynomial.

We will distinguish between *canonical* and *factored* polynomial representations of a Bayesian network. The canonical representation is unique, has exponential size, but is all we need to discuss the semantics of polynomial representations and their partial derivatives. Section 4 will then concern itself with computing factored polynomials, which size is determined by the network topology.

We start with some key notation first. Let $\mathbf{F} = \{X\} \cup \mathbf{U}$ be the family of variable $X$ and let $\mathbf{f} = x\mathbf{u}$ be a corresponding instantiation. We will then use $\theta_\mathbf{f}$ and/or $\theta_{x\mathbf{u}}$ to represent the conditional probability $Pr(x \mid \mathbf{u})$. Moreover, we will write $\mathbf{x} \sim \mathbf{f}$ to mean that instantiations $\mathbf{x}$ and $\mathbf{f}$ are consistent.

**Definition 1** *Let $\mathcal{N}$ be a Bayesian network with variables $\mathbf{X} = X_1, \ldots, X_n$ and families $\mathbf{F}_1, \ldots, \mathbf{F}_n$. Then*

$$\mathcal{F}(\lambda_{x_i}, \theta_{\mathbf{f}_i}) = \sum_{\mathbf{x}} \prod_{\mathbf{f}_i \sim \mathbf{x}} \theta_{\mathbf{f}_i} \prod_{x_i \sim \mathbf{x}} \lambda_{x_i}$$

*is called the canonical polynomial of network $\mathcal{N}$, where variables $\lambda_{x_i}$ are called evidence indicators and variables $\theta_{\mathbf{f}_i}$ are called network parameters.*

As an example, the canonical polynomial of the network in Figure 5 is:

$$\mathcal{F} = \theta_a \theta_{ab} \lambda_a \lambda_b + \theta_a \theta_{a\bar{b}} \lambda_a \lambda_{\bar{b}} + \theta_{\bar{a}} \theta_{\bar{a}b} \lambda_{\bar{a}} \lambda_b + \theta_{\bar{a}} \theta_{\bar{a}\bar{b}} \lambda_{\bar{a}} \lambda_{\bar{b}}.$$

In Definition 1, the instantiation $\mathbf{x}$ ranges over $ab, a\bar{b}, \bar{a}b, \bar{a}\bar{b}$; the instantiation $\mathbf{f}_i$ ranges over $a, \bar{a}, ab, a\bar{b}, \bar{a}b, \bar{a}\bar{b}$; and the instantiation $x_i$ ranges over $a, \bar{a}, b, \bar{b}$.

Clearly, the size of a canonical polynomial is exponential in the number of network variables. Moreover, the polynomial is independent of its quantification: two networks with the same structure have the same canonical polynomial. We formalize the notion of a quantification below.

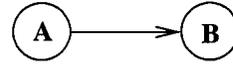

| A | $\phi_A$ |
|---|---|
| true | .3 |
| false | .7 |

| A | B | $\phi_B$ |
|---|---|---|
| true | true | .1 |
| true | false | .9 |
| false | true | .8 |
| false | false | .2 |

Figure 5: A Bayesian network. Table $\phi_A$ provides the prior probability of variable $A$ and Table $\phi_B$ provides the conditional probability of $B$ given $A$.

**Definition 2** *A quantification $\Theta$ of a Bayesian network is a function that assigns a value $\Theta(\mathbf{f})$ to each instantiation $\mathbf{f}$ of family $\mathbf{F}$.*

For example, according to the quantification $\Theta$ in Figure 5, we have $\Theta(ab) = .1$ and $\Theta(\bar{a}) = .7$.

The probability distribution represented by a Bayesian network is completely recoverable from its canonical polynomial:

**Definition 3** *The value of indicator $\lambda_x$ at instantiation $\mathbf{e}$, denoted $\mathbf{e}(x)$, is 1 if $x$ is consistent with $\mathbf{e}$, and is 0 otherwise. The value of polynomial $\mathcal{F}$ under evidence $\mathbf{e}$ and quantification $\Theta$ is defined as:*

$$\mathcal{F}(\mathbf{e}, \Theta) \stackrel{def}{=} \mathcal{F}(\lambda_{x_i} = \mathbf{e}(x_i), \theta_{\mathbf{f}_i} = \Theta(\mathbf{f}_i)).$$

Consider the canonical polynomial

$$\mathcal{F} = \theta_a \theta_{ab} \lambda_a \lambda_b + \theta_a \theta_{a\bar{b}} \lambda_a \lambda_{\bar{b}} + \theta_{\bar{a}} \theta_{\bar{a}b} \lambda_{\bar{a}} \lambda_b + \theta_{\bar{a}} \theta_{\bar{a}\bar{b}} \lambda_{\bar{a}} \lambda_{\bar{b}}$$

of the network in Figure 5. If the evidence $\mathbf{e}$ is $a\bar{b}$ and the quantification $\Theta$ is as given in Figure 5, then $\mathcal{F}(\mathbf{e}, \Theta)$ stands for $\mathcal{F}(\lambda_a = 1, \lambda_{\bar{a}} = 0, \lambda_b = 0, \lambda_{\bar{b}} = 1, \theta_a = .3, \theta_{\bar{a}} = .7, \theta_{ab} = .1, \theta_{a\bar{b}} = .9, \theta_{\bar{a}b} = .8, \theta_{\bar{a}\bar{b}} = .2)$, which equals to .27 in this case.

**Theorem 1** *Let $\mathcal{N}$ be a Bayesian network specifying distribution $Pr$ and having canonical polynomial $\mathcal{F}$. Then $\mathcal{F}(\mathbf{e}, \Theta) = Pr(\mathbf{e} \mid \Theta)$ for every evidence $\mathbf{e}$ and quantification $\Theta$.*

A canonical polynomial is then a function of many variables, where each variable corresponds to either an evidence indicator or a network parameter. For each instantiation $\mathbf{e}$ and quantification $\Theta$, the polynomial can be evaluated to compute the probability of $\mathbf{e}$ given $\Theta$. We will often write $\mathcal{F}(\mathbf{e})$ instead of $\mathcal{F}(\mathbf{e}, \Theta)$ when no ambiguity is anticipated.

Note that the canonical polynomial $\mathcal{F}$ of a Bayesian network is a multilinear function (i.e., each of its variables has degree one). Moreover, a monomial in $\mathcal{F}$ cannot contain two indicators of the same variable; neither can it contain two parameters of the same family. These properties are quite important as they imply that $\mathcal{F}$ is a linear function in the indicators of any variable; and is a linear function in the parameters of any family.

It is already observed in [19] that $Pr(\mathbf{e})$ is a linear function in each network parameter. More generally, it is shown in [1, 2] that $Pr(\mathbf{e})$ can be expressed as a polynomial of network parameters in which each parameter has degree one. The polynomials discussed in [1, 2], however, are constructed for a given evidence $\mathbf{e}$: they only contain one type of variables corresponding to network parameters, with no variables corresponding to evidence indicators. In fact, these polynomials correspond to our canonical polynomials when evidence indicators are fixed to a particular value. Hence, they do not represent network compilations as they cannot be used to answer queries with respect to varying evidence. Note also that the size of such polynomials is always exponential in the number of network variables, but we show in Section 4 how to factor them using the technique of variable elimination. The main insight we bring into such polynomial representations, however, is our results on the probabilistic semantics (and computational complexity) of their partial derivatives, a topic which we discuss next.

## 3 Probabilistic Semantics of Partial Derivatives

Our goal in this section is to show the probabilistic semantics associated with the partial derivatives of a network polynomial. Given these semantics, we then enumerate the class of probabilistic queries that can be answered once these derivatives are computed.

We need the following key notation first. Let $\mathbf{e}$ be an instantiation and $\mathbf{X}$ be a set of variables. Then $\mathbf{e}-\mathbf{X}$ denotes the subset of instantiation $\mathbf{e}$ pertaining to variables not appearing in $\mathbf{X}$. For example, if $\mathbf{e} = ab\bar{c}$, then $\mathbf{e} - A = b\bar{c}$ and $\mathbf{e} - AC = b$.

We start with the semantics of first derivatives:

**Theorem 2 (Semantics of 1st PDs)** *Let $\mathcal{N}$ be a Bayesian network representing probability distribution $Pr$ and let $\mathcal{F}$ be its canonical polynomial. For every variable $X$, family $\mathbf{F}$ and evidence $\mathbf{e}$:*

$$\partial \mathcal{F}(\mathbf{e}, \Theta)/\partial \lambda_x = Pr(x, \mathbf{e} - X | \Theta)$$

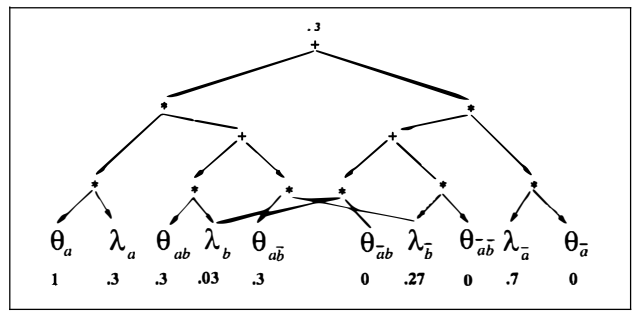

Figure 6: Partial derivatives are computed under $(\lambda_a, \lambda_{\bar{a}}, \lambda_b, \lambda_{\bar{b}}) = (1, 0, 1, 1)$ and $(\theta_a, \theta_{\bar{a}}, \theta_{ab}, \theta_{a\bar{b}}, \theta_{\bar{a}b}, \theta_{\bar{a}\bar{b}}) = (.3, .7, .1, .9, .8, .2)$.

$$\partial \mathcal{F}(\mathbf{e}, \Theta)/\partial \theta_{\mathbf{f}} = Pr(\mathbf{f}, \mathbf{e}|\Theta)/\Theta(\mathbf{f}).$$

That is, if we differentiate the polynomial $\mathcal{F}$ with respect to indicator $\lambda_x$ and evaluate the result at evidence $\mathbf{e}$, we obtain the probability of instantiation $x, \mathbf{e} - X$. A similar interpretation is given to the derivative of $\mathcal{F}$ with respect to parameter $\theta_{\mathbf{f}}$. Note that the second identity of Theorem 2 has indirectly been shown in [19], given that $\mathcal{F}(\mathbf{e}) = Pr(\mathbf{e})$.

Figure 6 depicts a factored polynomial representation of the Bayesian network in Figure 5, evaluated under evidence $a$, which we shall use as a running example. The partial derivatives of $\mathcal{F}$ with respect to each network indicator and parameter is shown below the indicator/parameter. For example, $\partial \mathcal{F}(a)/\partial \theta_a = 1$ and $\partial \mathcal{F}(a)/\theta_{\bar{a}} = 0$.

We now discuss some classical queries that can be answered based on first partial derivatives.[3]

The first class of queries relates to "what-if" analysis. Specifically, after having computed the probability of some evidence $\mathbf{e}$, in which variable $X$ is instantiated to some value $x'$, we ask: what would be the probability of $\mathbf{e}$ if the evidence on $X$ was different, say, $x$. According to Theorem 2, the answer to this query is exactly $\partial \mathcal{F}(\mathbf{e})/\partial \lambda_x$. In Figure 6 for example, where $\mathbf{e} = a$, suppose we ask: what would be the probability of evidence if $A$ was instantiated to $\bar{a}$ instead. That would be $\partial \mathcal{F}(\mathbf{e})/\partial \lambda_{\bar{a}} = .7$, which is also the prior probability of $\bar{a}$.

---

[3]Recovering probabilistic quantities from partial derivatives seems to be a standard technique in certain statistical literatures. There, a probability distribution is represented using its *generating function,* allowing one to recover means and variances from the partial derivatives of such a function. The main attraction of this approach is that the generating function tends to have a closed form, allowing the derivatives to have closed forms too—see [8] for an overview of the technique and its benefits.

Another class of queries that is immediately obtainable from partial derivatives is posterior marginals:

**Corollary 1 (Posterior Marginals)** *For $X \notin \mathbf{E}$:*

$$Pr(x \mid \mathbf{e}) = \frac{\partial \mathcal{F}(\mathbf{e})/\partial \lambda_x}{\mathcal{F}(\mathbf{e})}.$$

Therefore, the partial derivatives will give us the posterior marginal of every variable in constant time. In Figure 6, where $\mathbf{e} = a$, $Pr(b \mid \mathbf{e}) = (\partial \mathcal{F}(\mathbf{e})/\partial \lambda_b)/\mathcal{F}(\mathbf{e}) = .03/.3 = .1$ and $Pr(\bar{b} \mid \mathbf{e}) = (\partial \mathcal{F}(\mathbf{e})/\partial \lambda_{\bar{b}})/\mathcal{F}(\mathbf{e}) = .27/.3 = .9$.

The ability to compute such posteriors efficiently is probably the key celebrated property of jointree algorithms [10, 11], as compared to variable-elimination algorithms [20, 7, 21]. The latter class of algorithms is much simpler except that they can only compute such posteriors by invoking themselves once for each network variable, leading to a complexity of $O(n^2 \exp(w))$. Jointree algorithms can do this in $O(n \exp(w))$, however, but at the expense of a more complicated algorithm. The algorithm we are presenting based on the computation of partial derivatives can also compute all such marginals in only $O(n \exp(w))$ and seems to represent a middle ground in this efficiency/simplicity tradeoff.

One of the main complications in Bayesian network inference relates to the update of probabilities after having retracted evidence. This seems to pose no difficulties in the presented framework. For example, we can compute the posterior marginal of every instantiated variable, after the evidence on that variable has been retracted immediately.

**Corollary 2 (Evidence Retraction)** *For every variable $X$ and evidence $\mathbf{e}$, we have:*

$$Pr(\mathbf{e} - X) = \sum_x \frac{\partial \mathcal{F}(\mathbf{e})}{\partial \lambda_x};$$

$$Pr(x \mid \mathbf{e} - X) = \frac{\partial \mathcal{F}(\mathbf{e})/\partial \lambda_x}{\sum_x \partial \mathcal{F}(\mathbf{e})/\partial \lambda_x}.$$

In Figure 6, where $\mathbf{e} = a$, $Pr(\mathbf{e} - A) = \partial \mathcal{F}(\mathbf{e})/\partial \lambda_a + \partial \mathcal{F}(\mathbf{e})/\partial \lambda_{\bar{a}} = 1$ and $Pr(\bar{a} \mid \mathbf{e} - A) = (\partial \mathcal{F}(\mathbf{e})/\partial \lambda_{\bar{a}})/(\partial \mathcal{F}(\mathbf{e})/\partial \lambda_a + \partial \mathcal{F}(\mathbf{e})/\partial \lambda_{\bar{a}}) = .7/1 = .7$.

The above computation is the basis of an investigation of model adequacy [4, Chapter 10] and is typically implemented in the jointree algorithm using the technique of *fast retraction,* which requires a modification to the standard propagation method in jointrees [4, Page 104]. As given by the above theorem, we get this computation for free once we have partial derivatives with respect to network indicators.

Note that once we have the partial derivatives $\partial \mathcal{F}/\partial \theta_{\mathbf{f}}$, we can implement the APN algorithm for learning network parameters as shown in [19]. As for implementing the EM algorithm, we need the posterior marginals over network families [16], which are easily obtainable from the partial derivatives since

$$Pr(\mathbf{f} \mid \mathbf{e}, \Theta) = \frac{\partial \mathcal{F}(\mathbf{e}, \Theta)/\partial \theta_{\mathbf{f}}}{\mathcal{F}(\mathbf{e}, \Theta)} \Theta(\mathbf{f}).$$

For example, in Figure 6, $Pr(b, a \mid a)$ is $\frac{\partial \mathcal{F}(a)/\partial \theta_{ab}}{\mathcal{F}(a)} \theta_{ab} = \frac{.3}{.3} . 1 = .1$.

We now turn to the semantics of second derivatives:

**Theorem 3 (Semantics of 2nd PDs)** *Let $\mathcal{N}$ be a Bayesian network representing probability distribution $Pr$ and let $\mathcal{F}$ be its canonical polynomial. For every pair of variables $X \neq Y$, pair of families $\mathbf{F}_1 \neq \mathbf{F}_2$, and evidence $\mathbf{e}$:*

$$\partial^2 \mathcal{F}(\mathbf{e}, \Theta)/\partial \lambda_x \partial \lambda_y = Pr(x, y, \mathbf{e} - XY \mid \Theta);$$
$$\partial^2 \mathcal{F}(\mathbf{e}, \Theta)/\partial \lambda_x \partial \theta_{\mathbf{f}_1} = Pr(x, \mathbf{f}_1, \mathbf{e} - X \mid \Theta)/\Theta(\mathbf{f}_1);$$
$$\partial^2 \mathcal{F}(\mathbf{e}, \Theta)/\partial \theta_{\mathbf{f}_1} \partial \theta_{\mathbf{f}_2} = Pr(\mathbf{f}_1, \mathbf{f}_2, \mathbf{e} \mid \Theta)/\Theta(\mathbf{f}_1)\Theta(\mathbf{f}_2).$$

For example, the first identity reads: evaluating the derivative $\partial^2 \mathcal{F}/\partial \lambda_x \partial \lambda_y$ at evidence $\mathbf{e}$ and quantification $\Theta$ gives the probability $Pr(x, y, \mathbf{e} - XY \mid \Theta)$.

The significance of Theorems 2 and 3 is two fold:

1. *They show us how to compute answers to classical probabilistic queries by differentiating the polynomial representation of a Bayesian network.* Therefore, if we have an efficient way to generate the polynomial representation and to differentiate it, then we also have an efficient way to perform probabilistic reasoning. This is the view we promote in this paper.

2. *They show us how to compute valuable partial derivatives using classical probabilistic quantities.* The partial derivatives of $Pr(\mathbf{e})$ and $Pr(y \mid \mathbf{e})$ are important when estimating Bayesian network parameters and when performing sensitivity analysis. The third identity of Theorem 3, for example, shows us how to compute the second partial derivative of $Pr(\mathbf{e})$ with respect to two network parameters, $\theta_{\mathbf{f}_1}$ and $\theta_{\mathbf{f}_2}$, using the joint probability over their corresponding families, $Pr(\mathbf{f}_1, \mathbf{f}_2, \mathbf{e})$.

We have to note, however, that expressing partial derivatives in terms of classical probabilistic quantities requires some conditions: $\Theta(\mathbf{f}), \Theta(\mathbf{f}_1)$ and $\Theta(\mathbf{f}_2)$ cannot be 0 in Theorems 2 and 3. In the context of jointree algorithms, this complication can possibly be addressed using *lazy propagation*—see [12]. There is no such concern, however, if partial derivatives are computed directly as we do in this paper.

Theorems 2 and 3 facilitate the derivation of results relating to sensitivity analysis. Here's one example:

**Theorem 4 (Sensitivity)** *Let $\mathcal{N}$ be a Bayesian network specifying distribution $Pr$ and let $\mathcal{F}$ be its polynomial representation. For variable $Y \notin \mathbf{E}$:*

$$\frac{\partial Pr(y \mid \mathbf{e})}{\partial \theta_{x\mathbf{u}}}$$
$$= \frac{1}{\mathcal{F}(\mathbf{e})^2} \left( \frac{\partial^2 \mathcal{F}(\mathbf{e})}{\partial \theta_{x\mathbf{u}} \partial \lambda_y} \mathcal{F}(\mathbf{e}) - \frac{\partial \mathcal{F}(\mathbf{e})}{\partial \theta_{x\mathbf{u}}} \frac{\partial \mathcal{F}(\mathbf{e})}{\partial \lambda_y} \right)$$
$$= \frac{Pr(y, x, \mathbf{u} \mid \mathbf{e}) - Pr(y \mid \mathbf{e}) Pr(x, \mathbf{u} \mid \mathbf{e})}{Pr(x \mid \mathbf{u})}.$$

This theorem provides an elegant answer to the most central question of sensitivity analysis in Bayesian networks, as it shows how we can compute the sensitivity of a conditional probability to a change in some network parameter. The theorem phrases this computation in terms of both partial derivatives and classical probabilistic quantities—the second part, however, can only be used when $Pr(x \mid \mathbf{u}) \neq 0$.

One has to note an important issue here relating to co-varying parameters. Let $X$ be a variable in a Bayesian network and let $\mathbf{U}$ be its parents. We must then have $\sum_x \Theta(x\mathbf{u}) = 1$ for fixed $\mathbf{u}$. Therefore, when one of the parameters $\theta_{x\mathbf{u}}$ changes its value from $\Theta(x\mathbf{u})$ to $\Theta'(x\mathbf{u})$, all related parameters must also change so that $\sum_x \Theta'(x\mathbf{u}) = 1$ is maintained.

It is common to assume that parameters $\theta_{x\mathbf{u}}$, for a fixed $\mathbf{u}$, are all linear functions of some meta parameter $\tau_X$. That is, $\theta_{x\mathbf{u}} = \alpha_x \tau_X + \beta_x$, where $\alpha_x$ and $\beta_x$ are constants and $\sum_x \alpha_x \tau_X + \beta_x = 1$. Then using the generalized chain rule of differential calculus:

$$\partial Pr(y \mid \mathbf{e}) / \partial \tau_X = \sum_x \alpha_x \frac{\partial Pr(y \mid \mathbf{e})}{\partial \theta_{x\mathbf{u}}}.$$

Consider now the following three problems relating to sensitivity analysis. We want to compute the derivative $\partial Pr(y \mid \mathbf{e}) / \partial \tau_X$ for

*Problem (1):* every $Y$ and every $\tau_X$;

*Problem (2):* some $Y$ and every $\tau_X$;

*Problem (3):* every $Y$ and some $\tau_X$.

Given Theorem 4, and given that we can compute all first and second partial derivatives in $O(n^2 \exp(w))$ time, it follows immediately that Problem (1) can be solved in $O(n^2 \exp(w))$ time. We also show in [18] that the second partial derivative $\partial^2 \mathcal{F}(\mathbf{e}) / \partial \theta_{x\mathbf{u}} \partial \lambda_y$ can be computed in $O(n \exp(w))$ time for every parameter $\theta_{x\mathbf{u}}$ and a fixed indicator $\lambda_y$, or for a fixed parameter $\theta_{x\mathbf{u}}$ and every indicator $\lambda_y$. Therefore, Problems (2) and (3) can each be solved in $O(n \exp(w))$ time.

There seems to be two approaches for computing the derivative $\partial Pr(y \mid \mathbf{e}) / \partial \theta_{x\mathbf{u}}$, which has been receiving increased attention recently due to its role in sensitivity analysis and the learning of network parameters. We have just presented one approach where we found a closed form for $\partial Pr(y \mid \mathbf{e}) / \partial \theta_{x\mathbf{u}}$, using both partial derivatives and classical probabilistic quantities. The other approach capitalizes on the observation that $Pr(y \mid \mathbf{e})$ has the form $(\alpha \theta_{x\mathbf{u}} + \beta) / (\gamma \theta_{x\mathbf{u}} + \delta)$ for some constants $\alpha, \beta, \gamma$ and $\delta$ [1]. According to this second approach, one tries to compute the values of these constants based on the given Bayesian network and then computes the derivative of $(\alpha \theta_{x\mathbf{u}} + \beta) / (\gamma \theta_{x\mathbf{u}} + \delta)$ with respect to $\theta_{x\mathbf{u}}$. See [12, 14] for an example of this approach, where it is shown how to compute such constants using a limited number of propagations in the context of a jointree algorithm.

We now present yet another example of results that are facilitated by the view we promote in this paper. That is, we consider one of the key questions that arise in real–world diagnostic applications. Suppose we are given some evidence $\mathbf{e}$ and some hypothesis $Y$. We propagate the evidence and find that $Pr(y \mid \mathbf{e}) > Pr(\bar{y} \mid \mathbf{e})$; that is, $y$ is the fault given evidence $\mathbf{e}$. An expert may conclude differently, that $\bar{y}$ is more probable in this case, hence, indicating a problem in the Bayesian network model. Assuming that the network structure is correct, a key question is: which network parameters should we tweak in order to correct the problem — that is, leading to $Pr(y \mid \mathbf{e}) \leq Pr(\bar{y} \mid \mathbf{e})$. There is typically more than one parameter to tweak, but the following theorem provides key insights into answering this question:

**Theorem 5 (Parameter Tweaking)** *Let $Y$ and $X$ be binary variables in a Bayesian network $\mathcal{N}$ with polynomial representation $\mathcal{F}$. Let $\Theta$ and $\Theta'$ be two quantifications of $\mathcal{N}$ which agree on all parameters except those for $x\mathbf{u}$ and $\bar{x}\mathbf{u}$. If $Y \notin \mathbf{E}$, then*

$Pr(y \mid \mathbf{e}, \Theta') \leq Pr(\bar{y} \mid \mathbf{e}, \Theta')$ iff

$\Theta(x\mathbf{u})\Theta(\bar{x}\mathbf{u})(G - H) \leq$
$[\Theta'(x\mathbf{u}) - \Theta(x\mathbf{u})][\Theta(x\mathbf{u})(I - J) + \Theta(\bar{x}\mathbf{u})(K - L)],$

where

$$\begin{aligned}
G &= \partial \mathcal{F}(\mathbf{e}, \Theta)/\partial \lambda_y \\
H &= \partial \mathcal{F}(\mathbf{e}, \Theta)/\partial \lambda_{\bar{y}} \\
I &= \Theta(\bar{x}\mathbf{u})\partial^2 \mathcal{F}(\mathbf{e}, \Theta)/\partial \lambda_y \partial \theta_{\bar{x}\mathbf{u}} \\
J &= \Theta(\bar{x}\mathbf{u})\partial^2 \mathcal{F}(\mathbf{e}, \Theta)/\partial \lambda_{\bar{y}} \partial \theta_{\bar{x}\mathbf{u}} \\
K &= \Theta(x\mathbf{u})\partial^2 \mathcal{F}(\mathbf{e}, \Theta)/\partial \lambda_{\bar{y}} \partial \theta_{x\mathbf{u}} \\
L &= \Theta(x\mathbf{u})\partial^2 \mathcal{F}(\mathbf{e}, \Theta)/\partial \lambda_y \partial \theta_{x\mathbf{u}}.
\end{aligned}$$

Let us call $\Theta$ the pre-tweak quantification and $\Theta'$ the post-tweak quantification. The above theorem is stating conditions on the pre-tweak quantification, and the amount of tweaking, which would ensure that a certain hypothesis ranking is achieved in the post-tweak quantification. That is, the theorem allows us to compute the change to parameters $\theta_{x\mathbf{u}}$ and $\theta_{\bar{x}\mathbf{u}}$, respectively, which ensures a particular ranking on the hypotheses $y$ and $\bar{y}$.

For an example application of Theorem 5, consider the network in Figure 5 where $Pr(b) = .59 > Pr(\bar{b}) = .41$. We want to compute the amount of change to parameter $\theta_a$, initially set to $\Theta(a) = .3$, of root node $A$ which is needed to ensure that $Pr(b) \leq Pr(\bar{b})$. Applying Theorem 5:

$$\Theta'(a) - \Theta(a) \geq \frac{(.3)(.7)(.59 - .41)}{.3(.56 - .14) + .7(.27 - .03)} = .12857.$$

That is, $\Theta'(a) \geq .42857$ ensures that $Pr(b) \leq Pr(\bar{b})$.

We close this section by a note on the complexity of applying Theorem 5. Since we can compute all first and second partial derivatives in $O(n^2 \exp(w))$ time, we are able to compute in $O(n^2 \exp(w))$ time, for each variable $Y$ and each variable $X$, the amount of change needed in the parameters of $X$ to reverse the probabilistic ranking of $Y$'s values. Moreover, we show in [18] that we can perform this computation in $O(n \exp(w))$ time only, for a fixed $Y$ or a fixed $X$.

## 4 Compiling a Bayesian Network

Our goal in this section is to present a variable-elimination algorithm for compiling a *factored* polynomial representation of a Bayesian network. This is to be contrasted with a *canonical* polynomial representation (Definition 1), which is straightforward to obtain but requires exponential space.

As an example, following is the canonical polynomial of the network in Figure 5:

$\mathcal{F} = \theta_a \theta_{ab} \lambda_a \lambda_b + \theta_a \theta_{a\bar{b}} \lambda_a \lambda_{\bar{b}} + \theta_{\bar{a}} \theta_{\bar{a}b} \lambda_{\bar{a}} \lambda_b + \theta_{\bar{a}} \theta_{\bar{a}\bar{b}} \lambda_{\bar{a}} \lambda_{\bar{b}},$

and following is one of its factored representations:

$\mathcal{F} = \theta_a \lambda_a(\theta_{ab}\lambda_b + \theta_{a\bar{b}}\lambda_{\bar{b}}) + \theta_{\bar{a}}\lambda_{\bar{a}}(\theta_{\bar{a}b}\lambda_b + \theta_{\bar{a}\bar{b}}\lambda_{\bar{b}}).$

The algorithm we present requires an ordering $\pi$ of the variables in a Bayesian network $\mathcal{N}$. The algorithm goes as follows:

1. For every variable $X$ with parents $\mathbf{U}$, *parameterize* the CPT of $X$, $\phi_X$, as follows. Replace each entry in $\phi_X$ which corresponds to instantiation $x\mathbf{u}$ by $\theta_{x\mathbf{u}}\lambda_x$.

2. Consider every variable $X$ according to order $\pi$:
   (a) Multiply all tables that contain variable $X$ to yield table $\phi$.
   (b) Replace the multiplied tables by the result of summing out $X$ from $\phi$, sum_out$(\phi, X)$.

3. Only one table $\phi$ will be left, with a single entry. Return the single entry of $\phi$.

We will call this algorithm VE_COMPILE$(\mathcal{N}, \pi)$.

Following is an illustrative example of VE_COMPILE: we will use the elimination order $\pi = <B, A>$ to compile the network in Figure 5. The parameterized CPTs are given below:

| A | $\phi_A$ |
|---|---|
| true | $\theta_a \lambda_a$ |
| false | $\theta_{\bar{a}} \lambda_{\bar{a}}$ |

| A | B | $\phi_B$ |
|---|---|---|
| true | true | $\theta_{ab}\lambda_b$ |
| true | false | $\theta_{a\bar{b}}\lambda_{\bar{b}}$ |
| false | true | $\theta_{\bar{a}b}\lambda_b$ |
| false | false | $\theta_{\bar{a}\bar{b}}\lambda_{\bar{b}}$ |

To eliminate variable $B$, we sum it out from table $\phi_B$, since it is the only table containing $B$:

| A | sum_out$(\phi_B, B)$ |
|---|---|
| true | $\theta_{ab}\lambda_b + \theta_{a\bar{b}}\lambda_{\bar{b}}$ |
| false | $\theta_{\bar{a}b}\lambda_b + \theta_{\bar{a}\bar{b}}\lambda_{\bar{b}}$ |

To eliminate variable $A$, we must multiply the above table with table $\phi_A$ and then sum-out variable $A$ from the result. Multiplying gives:

| A | $\phi_A$sum_out$(\phi_B, B)$ |
|---|---|
| true | $\theta_a \lambda_a(\theta_{ab}\lambda_b + \theta_{a\bar{b}}\lambda_{\bar{b}})$ |
| false | $\theta_{\bar{a}} \lambda_{\bar{a}}(\theta_{\bar{a}b}\lambda_b + \theta_{\bar{a}\bar{b}}\lambda_{\bar{b}})$ |

Summing out, we get

| sum_out$(\phi_A$sum_out$(\phi_B, B), A)$ |
|---|
| $\theta_a \lambda_a(\theta_{ab}\lambda_b + \theta_{a\bar{b}}\lambda_{\bar{b}}) + \theta_{\bar{a}} \lambda_{\bar{a}}(\theta_{\bar{a}b}\lambda_b + \theta_{\bar{a}\bar{b}}\lambda_{\bar{b}})$ |

The polynomial,

$$\mathcal{F}(\lambda_a, \lambda_{\bar{a}}, \lambda_b, \lambda_{\bar{b}}, \theta_a, \theta_{\bar{a}}, \theta_{ab}, \theta_{a\bar{b}}, \theta_{\bar{a}b}, \theta_{\bar{a}\bar{b}})$$
$$= \theta_a \lambda_a (\theta_{ab} \lambda_b + \theta_{a\bar{b}} \lambda_{\bar{b}}) + \theta_{\bar{a}} \lambda_{\bar{a}} (\theta_{\bar{a}b} \lambda_b + \theta_{\bar{a}\bar{b}} \lambda_{\bar{b}})$$

is then the result of this compilation and is depicted in Figure 6 as a rooted DAG. When the parameters in such a compilation are fixed to some numeric values, the result is still a polynomial and is a special case of a Query-DAG [6]. Note, however, that a Query-DAG can have multiple roots corresponding to different queries which must be specified before the compilation process takes place. The compilation as presented above, however, is only meant to answer one query: computing the probability of given evidence. But as we have shown earlier, many other queries can be computed from the partial derivatives of this compilation.

We can now state the soundness/complexity result:

**Theorem 6** *Given a Bayesian network $\mathcal{N}$ and a corresponding variable elimination order $\pi$ of width $w$ and length $n$, the call VE_COMPILE($\mathcal{N}, \pi$) returns a polynomial representation of network $\mathcal{N}$ of size $O(n \exp(w))$ and in $O(n \exp(w))$ time.*

## 5 Evaluating and Differentiating a Polynomial Representation

We now turn to the process of evaluating a network compilation $\mathcal{F}$ under some evidence $\mathbf{e}$ and quantification $\Theta$, and then computing all its partial derivatives. We will first describe the computation procedurally using a message–passing scheme and then provide the semantics of such computation.

As it turns out, the evaluation of $\mathcal{F}$ and computing its partial derivatives can be accomplished using a two-phase message passing scheme in which each message is simply a number. In the first phase, messages are sent from nodes to their parents in $\mathcal{F}$, leading to an evaluation of $\mathcal{F}$. In the second phase, messages are sent from nodes to their children, leading to the computation of all partial derivatives.

We will associate two attributes with each node i in the compilation $\mathcal{F}$: a value val(i) and a partial derivative pd(i). The goal of message passing is to set the value of each of these attributes under some evidence $\mathbf{e}$ and quantification $\Theta$. We are mainly interested in the value of root node r, val(r), as it represents $\mathcal{F}(\mathbf{e}, \Theta)$, and the partial derivative of each leaf node l, pd(l), at it represents $\partial \mathcal{F}(\mathbf{e}, \Theta) / \partial l$. Message passing proceeds in two phases: the first phase

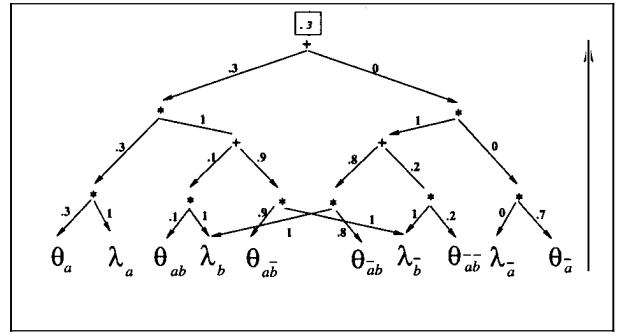

Figure 7: Passing val-messages under evidence $\mathbf{e} = a$ and quantification $\Theta : (\theta_a, \theta_{\bar{a}}, \theta_{ab}, \theta_{a\bar{b}}, \theta_{\bar{a}b}, \theta_{\bar{a}\bar{b}}) = (.3, .7, .1, .9, .8, .2)$.

sets the value of each node, and the second phase sets the partial derivatives.

**Passing val-messages** In the first phase, a val-message is passed from each node to all its parents according to the following rules:

1. *Initiation:* Each <u>leaf node</u> l sets its value:
   - val(l) ← $\mathbf{e}(x)$ when l is an indicator $\lambda_x$;
   - val(l) ← $\Theta(x\mathbf{u})$ when l is a parameter $\theta_{x\mathbf{u}}$.
   
   Node l then sends a message mes(l, k) = val(l) to each parent k.

2. *Iteration I:* When an <u>addition node</u> i receives messages from all its children j, it sets its value val(i) ← $\sum_j$ mes(j, i) and then sends a message mes(i, k) = val(i) to each parent k.

3. *Iteration II:* When a <u>multiplication node</u> i receives messages from <u>all its children j</u>, it sets its value val(i) ← $\prod_j$ mes(j, i) and then sends a message mes(i, k) = val(i) to each parent k.

4. *Termination:* The first phase terminates when the <u>root node</u> r receives messages from all its children. In that case, $\mathcal{F}(\mathbf{e}, \Theta)$ = val(r).

This process is illustrated in Figure 7, where it leads to assigning val(r) = .3, indicating that the probability of evidence is $\mathcal{F}(\mathbf{e}, \Theta) = .3$.

**Passing pd-messages** In the second phase, a pd-message is passed from each node to all its children:

1. *Initiation:* The <u>root node</u> r sets pd(r) ← 1 and sends message mes(r, j) = 1 to each child j.

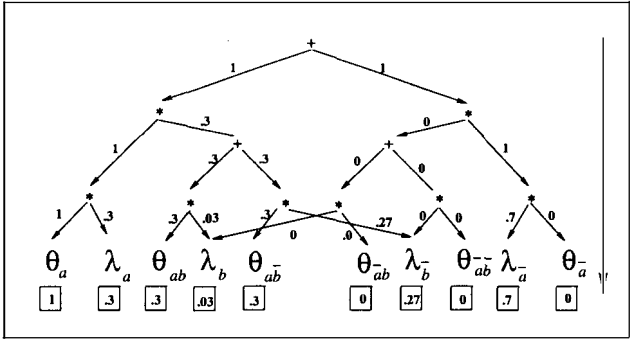

Figure 8: Passing pd-messages under evidence $\mathbf{e} = a$ and quantification $\Theta : (\theta_a, \theta_{\bar{a}}, \theta_{ab}, \theta_{a\bar{b}}, \theta_{\bar{a}b}, \theta_{\bar{a}\bar{b}}) = (.3, .7, .1, .9, .8, .2)$.

2. *Iteration I:* When an addition node i receives messages from all its parents k, it sets $\mathsf{pd}(i) \leftarrow \sum_k \mathsf{mes}(k, i)$ and then sends a message $\mathsf{mes}(i, j) = \mathsf{pd}(i)$ to each child j.

3. *Iteration II:* When a multiplication node i receives messages from all its parents k, it sets $\mathsf{pd}(i) \leftarrow \sum_k \mathsf{mes}(k, i)$ and then sends a message $\mathsf{mes}(i, j) = \mathsf{pd}(i) \prod_{j' \neq j} \mathsf{val}(j')$ to each child j, where j' is a child of i.

4. *Termination:* The second phase terminates when each leaf node l receives messages from all its parents. In that case, $\mathsf{pd}(l)$ is guaranteed to be the partial derivative $\partial \mathcal{F}(\mathbf{e}, \Theta)/\partial l$.

This process is illustrated in Figure 8, assuming that val-messages have been propagated as in Figure 7. The process leads to assigning pd(l) to each leaf node in the compilation, shown in a box below the leaf node. For example, $\partial \mathcal{F}(\mathbf{e}, \Theta)/\partial \lambda_b = .03$.

Clearly, the number of messages passed in both phases is twice the number of edges in the compilation $\mathcal{F}$. And since each message can be computed in constant time, the whole computation is linear in the size of $\mathcal{F}$. We are currently working on counting the number of operations performed by our message passing scheme and comparing that to other frameworks for multiple-query inference, such as the HUGIN and Shenoy/Shafer architectures.

**The Semantics of message passing** We now turn to the semantics of our message-passing scheme. Specifically, any node i in function $\mathcal{F}$ can be viewed as a subfunction $\mathcal{F}_i$ of $\mathcal{F}$. For any evidence $\mathbf{e}$, our message passing scheme guarantees that $\mathsf{val}(i) = \mathcal{F}_i(\mathbf{e}, \Theta)$. Alternatively, node i can be viewed as an intermediate variable $V_i$ of function $\mathcal{F}$. Our message passing scheme guarantees that $\mathsf{pd}(i) = \partial \mathcal{F}(\mathbf{e}, \Theta)/\partial V_i$.

These results follow directly from the work of [9], which addresses the more general problem of computing partial derivatives for a function that has an *acyclic computation graph*. Our network compilations $\mathcal{F}$ fall into that category.

We close this section by two remarks. The first regards the computation of rounding errors when evaluating $\mathcal{F}(\mathbf{e}, \Theta)$ using a limited-precision computer. Specifically, let $\delta i$ be the rounding error "generated" when computing the value of node i—that is, the difference between the value computed using limited-precision and infinite-precision. If $|\delta i| \leq \epsilon |\mathsf{val}(i)|$ for a machine–specific $\epsilon$, then the rounding error in computing $\mathcal{F}(\mathbf{e}, \Theta)$, $\Delta \mathcal{F}(\mathbf{e}, \Theta)$, can be bound as follows (see [17] for details):

$$\Delta \mathcal{F}(\mathbf{e}, \Theta) \leq \epsilon \sum_{\text{non-leaf } i} |\mathsf{pd}(i)\mathsf{val}(i)|.$$

Note that bounding the rounding error can be done simultaneously during the passage of pd-messages, and requires no extra space.

Our final remark is on the computation of second partial derivatives. The partial derivative $\mathsf{pd}(i)$ is by itself a function of many variables and can be differentiated again using the same approach. In particular, if the compilation $\mathcal{F}$ has $O(n)$ variables, then its second partial derivatives can be computed simultaneously in $O(n |\mathcal{F}|)$ time and space, where $|\mathcal{F}|$ is the size of $\mathcal{F}$. Again, this follows from a more general result in [17].

Therefore, if the compilation $\mathcal{F}$ of a Bayesian network of size $n$ is induced using an elimination order of width $w$, as suggested earlier, then the value of polynomial $\mathcal{F}$; all its first partial derivatives; and all its second partial derivatives can be computed simultaneously in $O(n^2 \exp(w))$ time and space. This is quite interesting given the number of queries that can be answered in constant time given such derivatives. We are unaware of any computational framework for Bayesian networks, which combines the simplicity, comprehensiveness and computational complexity of the presented framework based on partial derivatives.

## 6 Conclusion

We have presented one of the simplest, yet most comprehensive frameworks for inference in Bayesian networks. According to this framework, one compiles a Bayesian network into a multivariate polynomial $\mathcal{F}$ using a variable elimination scheme. For each piece of evidence e, one then traverses the polynomial $\mathcal{F}$ twice. In the first pass, the value of $\mathcal{F}$ is computed. In the second pass, all its partial derivatives are computed. Once such derivatives are made available, one can compute in constant time answers to a very large class of probabilistic queries, relating to classical inference, parameter estimation, model validation and sensitivity analysis.

The proposed framework makes two key contributions to the state-of-the-art on probabilistic reasoning. First, it highlights a key and comprehensive role of partial differentiation in this form of reasoning, shedding new light on its utility in sensitivity analysis. Second, it presents one of the simplest, yet most comprehensive frameworks for inference in Bayesian networks, which lends itself to cost-effective implementations on a variety of software and hardware platforms. Note that compiling a Bayesian network can be performed off-line. The only computation that is needed on-line is that of passing val-messages and pd-messages, which is quite simple. We are currently investigating the development of dedicated hardware for implementing the proposed message passing scheme. Similar dedicated hardware has been developed for other reasoning frameworks, such as genetic programming and neural networks, and can provide a valuable tool for computationally intensive tasks in Bayesian networks, such as the estimation of network parameters using APN/EM.

The development of such hardware is also expected to help in migrating Bayesian-network applications to embedded systems — such as consumer electronics — which are characterized by their primitive computational resources, therefore, affording to host only the very simplest reasoning frameworks.